\documentclass[11pt]{article}
\usepackage[utf8]{inputenc}
	\usepackage[utf8]{inputenc}
	\usepackage[english]{babel}
	\usepackage{amsfonts}
	\usepackage{amssymb}
	\usepackage{enumitem}
	\usepackage{bbm}
	\usepackage{mathtools}
	\usepackage{subcaption}
	\usepackage{booktabs}	
	\usepackage{setspace}
	\usepackage{graphicx}
	\usepackage{fullpage}
	
	\usepackage{soul}
	\usepackage{floatrow}
	\usepackage{placeins}
	\usepackage{siunitx}
	
\newcommand\blfootnote[1]{%
  \begingroup
  \renewcommand\thefootnote{}\footnote{#1}%
  \addtocounter{footnote}{-1}%
  \endgroup
}

	\usepackage{natbib}
	\usepackage{hyperref}

	\renewcommand{\P}{\mathop{}\!\textnormal{P}}
	\newcommand{\E}{\mathop{}\!\textnormal{E}}

	\newcommand{\Var}{\mathop{}\!\textnormal{Var}}
	
	\newcommand{\N}{\mathcal{N}}

	\newcommand{\0}{\mathbf{0}}

	\newcommand{\R}{\mathbb{R}}

	\newcommand{\fh}{\ensuremath{\hat{f}}}  % Empirical prediction

    \usepackage{amsthm}
    \usepackage{thmtools}
    \usepackage{xparse}

        \makeatletter
    \newtheoremstyle{indented}
      {20pt}% space before
      {20pt}% space after
      {\addtolength{\@totalleftmargin}{1.5em}
       \addtolength{\linewidth}{-1.5em}
       \parshape 1 1.5em \linewidth}% body font
      {}% indent
      {\bfseries}% header font
      {.}% punctuation
      {.5em}% after theorem header
      {}% header specification (empty for default)
    \makeatother

    \newcommand{\newthm}[2]{
        \theoremstyle{indented}
        \newtheorem{#1}[basethm]{#2}
        \expandafter\def\csname #1autorefname\endcsname{#2}
    }
    \newcommand{\newannot}[2]{
        \theoremstyle{indented}
        \newtheorem{#1}{#2}[annotthm]
        \expandafter\def\csname #1autorefname\endcsname{#2}
    }
    \newcommand{\newthmplus}[2]{
        \theoremstyle{indented}
        \newtheorem{#1xxx}{#2}
        \expandafter\def\csname #1autorefname\endcsname{#2}
        
        \NewDocumentEnvironment{#1*}{o d()}
            {
                \IfNoValueF{##2}{
                    \expandafter\def\csname the#1xxx\endcsname{##2}
                    \addtocounter{basethm}{-1}
                }
                \IfNoValueTF{##1}{\begin{#1xxx}}{\begin{#1xxx}[##1]}
                    \IfNoValueF{##2}{\label{#1:##2}}
            }
            {
                \end{#1xxx}
                
                \IfNoValueTF{##2}{\global\let\theannotthm\thebasethm}{\global\def\theannotthm{##2}}
            }
        
        \NewDocumentEnvironment{#1}{o d()}
            {
                \IfNoValueTF{##1}{\begin{#1xxx}}{\begin{#1xxx}[##1]}
                    \IfNoValueF{##2}{\label{#1:##2}}
            }
            {
                \end{#1xxx}
                \global\let\theannotthm\thebasethm
            }
    }

    \newthmplus{theo}{Theorem}

    \newthmplus{defi}{Definition}
    \newthmplus{asm}{Assumption}
    \newthmplus{lem}{Lemma}
    \newthmplus{prop}{Proposition}
    \newthmplus{rem}{Remark}
    \newthmplus{res}{Result}
    \newthmplus{cor}{Corollary}
    \newthmplus{ex}{Example}
    \newthmplus{con}{Conjecture}
    \newthmplus{clm}{Claim}
    \newthmplus{stp}{Setup}
    \newthmplus{simu}{Simulation}
    \newthmplus{idea}{Idea}
    \newthmplus{ins}{Insight}
    \newthmplus{ques}{Question}
    \newthmplus{cnp}{Concept}
    
    \newannot{att}{Attribution}
    \newannot{com}{Comment}
    \newannot{note}{Note}
    \newannot{thou}{Thought}
    \newannot{ext}{Extension}
    \newannot{apl}{Application}

\title{
    Machine-Learning Tests for Effects on Multiple Outcomes
}
\author{
    Jens Ludwig
    \and Sendhil Mullainathan
    \and Jann Spiess
}
\date{This draft: May 9, 2019}

\begin{document}

    \maketitle%
    \blfootnote{
        Jens Ludwig, Harris School of Public Policy, University of Chicago and NBER, \href{mailto:jludwig@uchicago.edu}{jludwig@uchicago.edu}.
        Sendhil Mullainathan, Booth School of Business, University of Chicago and NBER, \href{mailto:sendhil.mullainathan@chicagobooth.edu}{sendhil.mullainathan@chicagobooth.edu}.
        Jann Spiess, Microsoft Research New England, \href{mailto:jspiess@stanford.edu}{jspiess@stanford.edu}.
        An earlier versions of this manuscript was titled ``Testing Effects on Groups of Outcomes'' (November 2016).
        We thank seminar participants at Harvard University, the University of Chicago, and the University of Pennsylvania for helpful comments.
    }%
    
    \vspace{-2em}
    
    \begin{abstract}
        In this paper we present tools for applied researchers that re-purpose off-the-shelf methods from the computer-science field of machine learning to create a ``discovery engine'' for data from randomized controlled trials (RCTs). The applied problem we seek to solve is that economists invest vast resources into carrying out RCTs, including the collection of a rich set of candidate outcome measures. But given concerns about inference in the presence of multiple testing, economists usually wind up exploring just a small subset of the hypotheses that the available data could be used to test. This prevents us from extracting as much information as possible from each RCT, which in turn impairs our ability to develop new theories or strengthen the design of policy interventions. Our proposed solution combines the basic intuition of reverse regression, where the dependent variable of interest now becomes treatment assignment itself, with methods from machine learning that use the data themselves to flexibly identify whether there is any function of the outcomes that predicts (or has signal about) treatment group status. This leads to correctly-sized tests with appropriate $p$-values, which also have the important virtue of being easy to implement in practice.
        One open challenge that remains with our work is how to meaningfully interpret the signal that these methods find.
    \end{abstract}

    \section{Introduction}
    
In this paper we present tools for applied researchers that re-purpose off-the-shelf methods from the computer-science field of machine learning to create a ``discovery engine'' for data from randomized controlled trials (RCTs). The applied problem we seek to solve is that economists invest vast resources into carrying out RCTs, including the collection of a rich set of candidate outcome measures. But given concerns about inference in the presence of multiple testing, economists usually wind up exploring just a small subset of the hypotheses that the available data could be used to test. This prevents us from extracting as much information as possible from each RCT, which in turn impairs our ability to develop new theories or strengthen the design of policy interventions.

As a concrete example, consider an RCT of some health-related intervention like subsidized health insurance. This type of RCT would typically include original in-person data collection to supplement administrative data like electronic health records (EHR). Since these types of health interventions could plausibly affect a wide range of health problems (or precursors to health problems) as well as health-related behaviors, we typically cast a wide net during the data-collection stage and assemble as wide a range of plausibly relevant measures as we can (recognizing this is just a subset of what could be potentially measured).

Then we reach the analysis stage. Because under most multiple-testing procedures the penalty to a given outcome’s $p$-value increases as the number of outcomes considered increases, all else equal we usually try to discipline ourselves and limit the number of measures we turn into outcomes to analyze. Then there is the question of how to turn measures into outcomes. Our theory says ``health'' should be affected, but does that mean self-reported global health or specific health problems or particular physical limitations or body mass index or cholesterol or glycated hemoglobin? Should we focus just on how the intervention affects the mean values of these outcomes, or should we explore other parts of the distributions for some outcomes? Or should we examine the joint distributions of multiple outcomes, for example if we thought ``self-reported health very good \textit{and} sees doctor for regular check-ups'' might be particularly affected and also reveal something about underlying mechanisms of action?

In the end, applied researchers do the best they can using some combination of theory, previous findings and intuition to specify which hypotheses to test. But as our example makes abundantly clear, this is usually just a small subset of the hypotheses that \textit{could} be tested. Of course, applied researchers know this better than anyone. They also know that exploration often leads to the biggest discoveries. For example, when Congress passed the Housing and Community Development Act of 1992 that set aside \$102 million for the Moving to Opportunity (MTO) demonstration, it required HUD to submit a report to Congress ``describing the long-term housing, employment, and educational achievements of the families assisted under the demonstration program,'' as well as ``any other information the Secretary considers appropriate in evaluating the demonstration.'' But qualitative interviews with the MTO participants raised the possibility that the biggest changes occured not in the areas of housing, employment, or education, but rather the reduced trauma and anxiety associated with substantial gains in safety \citep{Kling:2007cl}. Future quantitative evaluations confirmed that some of the most important impacts fell under what at the launch of MTO were thought to be either so unimportant or unlikely that they were relegated to the category of ``any other information'' \citep{Kling:2007cl,sanbonm2011,ludwig2011,ludwig2012,kessler2014}.

The machine-learning approach we propose here is a ``discovery engine'' intended to complement existing joint- and multiple-testing methods that are based on ex-ante curation by the researcher. It is based on two simple observations from the statistics and computer-science literature. First, the `false-positive' problem that a group of outcomes poses can be thought of finding something in a given experiment that is idiosyncratic to that sample, rather than a true feature of the underlying data-generating process. Put this way, we can see that the general structure of this concern is similar to the concern within machine learning of `over-fitting,' which has been the focus of a large literature in statistics and computer science. We can thus leverage sample-splitting methods from the standard machine-learning playbook, which are designed to control over-fitting to ensure that statistical models capture true structure in the world rather than idiosyncracies of any particular dataset.

The second simple observation, going back to at least \cite{friedman2004multivariate}, is that the question of whether treatment $T$ (say binary) affects a group of variables $Y= (Y_1,...,Y_k)$ in an experiment is equivalent to the question whether $T$ is predictable using $Y$ (better than some trivial benchmark). In the parlance of \cite{Kleinberg:2015eo} and \cite{Mullain:2017}, this simple observation turns testing effects on many outcome variables into a prediction task (``$\hat{y}$ problem'', which here really is a ``$\hat{T}$ problem''). Predictability takes the place of a treatment effect on the full distribution of outcomes. This formulation allows us to leverage data-driven predictors from the machine-learning literature to flexibly mine for effects, rather than rely on more rigid approaches like multiple-testing corrections and pre-analysis plans.
    
We show that, for any samples size, this test produces a $p$-value that is exactly sized under the null hypothesis of jointly no effect on the group of outcomes. We also discuss how we can use our procedure to learn something about {\em where} any effect happens for purposes like testing theories or carrying out benefit--cost analyses of specific interventions.
In ongoing work, we also extend the test to deal with typical features of real-world experiments, namely missing data, conditional or clustered randomization, and stratification. And since this method is based on re-purposing existing off-the-shelf methods from machine learning, it has (from the perspective of applied research) the great virtue of being quite straightforward to implement.

In framing the many-outcomes problem as testing whether two distributions are the same, we relate our work to the general two-sample problem, in particular  non-parametric approaches based on matching \citep[such as][]{rosenbaum2005exact} or on kernels \citep[e.g.][]{gretton2007kernel}. In terms of using a prediction approach to the two-sample problem, we are building on a classic literature in economics and econometrics that studies discrimination using reverse regressions \citep[as discussed e.g. by][]{Goldberger:1984bp}, as well as a literature in statistics that connects the two-sample problem to classification \citep[going back to][]{friedman2004multivariate}. Relatedly, \cite{Gagnon-Bartsch2019-ay} develop a classification test for differences between two distributions, focusing on testing covariate imbalance in experiments. Like them, we use a permutation test to obtain a valid $p$-value, paralleling recent uses of omnibus permutation tests and randomization inference in the evaluation of experiments \citep{Potter:2006kt,Ding:2015hza,Chetty:2016ev,Young:2016ve}.

In terms of typical applications, our research is related to multiple testing procedures based on individual mean comparison tests.
Multiple testing procedures control the overall probability of at least one false positive (family-wise error rate) or the proportions of false positives among all rejections \citep[false discovery rate,][]{benjamini1995controlling}.
The most prominent such procedure, the Bonferroni correction \citep{bonferroni1936teoria,dunn1961multiple}, and its \cite{holm1979simple} step-wise improvement, ignore the correlation between individual test statistics.
Other, more recent procedures take the dependence structure of test statistics into account, for example the step-wise procedure by \cite{Romano:2005bw};
most closely related to our framework, 
\cite{List:2016dt} propose a bootstrap-based adaption to experiments.

We set up the many-outcomes problem in Section~\ref{sect:setup}. In Section~\ref{sect:standardapproaches}, we review some standard approaches, and discuss their applicability in selected economic examples. Section~\ref{sect:predicton} presents our prediction procedure and analyzes its properties. In Section~\ref{sect:analyze}, we present some ideas on interpreting the prediction output.
Section~\ref{sect:difference} proposes a framework and concrete tools for directly providing simple representations of the causal effect on the distribution of outcomes.
Section~\ref{sect:simulation} illustrates the testing approach on simulated data.
In Section~\ref{sect:conclusion}, we conclude by discussing challenges from typical features of real-world experiments.

    \section{Setup}
    \label{sect:setup}
    
    We consider the problem of testing whether a group of outcomes variables is affected in an experiment.
    Assume we have a sample of $n$ iid observations $(T_i,Y_i)$, with binary treatment $T_i$ assigned randomly. Randomization may be by clusters of observation and within some strata.     
    \begin{align*}
                    S = \{ (T_1,Y_1),\ldots,(T_n,Y_n) \}
                \end{align*}
            
    We have a group of $k$ scalar outcomes: 
    \begin{align*}
        Y_i=(Y_{i1},\ldots,Y_{ik}).
    \end{align*}
    Our primary goal is to test whether treatment has an effect on this group of outcomes, that is, whether the distributions of $Y|T=1$ is the same as the distribution of $Y|T=0$.
    If, for example, the two distributions differ only by a ($k$-dimensional) mean shift by $\tau = (\tau_1,\ldots,\tau_k)'$, our null hypothesis of interest is $\tau_j = 0$ for all $j$ simultaneously.
    
    A related but different question is whether there are effects on the \textit{individual} outcome variables $Y_{ij}$. This question can be useful for testing a specific theory, or if researchers have a clear sense of which individual outcome variable is of central interest. But often this question is complementary to the testing the {\em joint} hypothesis of no effect on any outcome. Having run a complex and expensive trial, a minimal first-pass question is commonly to ask the question ``Did the experiment have any effect at all?'' Tests of this joint hypothesis can be useful even when researchers are interested in a single specific hypothesis, for example when we do not know how to convert a specific hypothesis into individual variables (such as ``health'' being measured via numerous outcomes). 
      
    \section{Standard Approaches and their Mapping to Applications}
    \label{sect:standardapproaches}

      In this section we consider the standard approaches to testing whether treatment has an effect on multiple outcomes. Standard approaches fall into two broad categories: 
      {\em Tests based on mean comparisons of the outcomes}, such as a Wald test in a seemingly unrelated regression (SUR) of the outcomes on the treatment dummy; and {\em tests based on a pre-defined index}, where we aggregate all outcomes (usually linearly) into a single index and run one test on the treatment-control mean difference for this index \citep[e.g.][web appendix]{Kling:2007cl}. We begin by considering the implicit assumptions behind these standard approaches. We then consider how these assumptions fare when confronted with the range of applications that experimental economists may encounter in practice.

    \subsection{Standard Testing Approaches}

        We first consider the assumptions behind tests of effects on a group of outcomes that are based on aggregating the results of individual mean-comparison tests. Before we consider the aggregation of mean values, it is useful to first focus on the means tests themselves.
        
        To simplify the exposition we consider a simple class of data-generating processes. For outcome $j$, write
            \begin{align*}
                Y_j = C_j + \tau_j T_j + \epsilon_j,
            \end{align*}
        where the control baseline $C_j$ is constant, while the treatment effect $\tau_j$ and the mean-zero error term $\epsilon_j$ are random and possibly correlated both within and between outcomes, but independent of treatment assignment $T_j$. For simplicity, we assume here that the $\tau_j$ and $\epsilon_j$ are jointly Normally distributed. Our goal is to test whether treatment and control distributions are the same -- that is, whether $\tau_j=0$ for all $j$.
        What would be an appropriate test for the null hypothesis?
        
        Many standard approaches are based on individual mean comparisons, that is, they estimate the individual mean differences and then aggregate those to test that all mean differences are simultaneously zero.
        If the variances were different between treatment and control groups, testing for $\tau_j=0$ by only testing means would potentially leave information about different spreads on the table, no matter how we aggregate between individual mean estimates. If we base our inference solely on mean comparisons, we thus implicitly assume that mainly the means are affected by the intervention.
        
        Individual means tests alone do not directly provide a rejection criterion for the overall null hypothesis of no effect on the full group of outcomes; to obtain the latter, we need to aggregate individual estimates while ensuring that our test is properly sized (that is, that the probability of rejecting the null hypothesis does not exceed the desired level when there is indeed no effect on the outcomes).
        A particular simple form of aggregation are \textit{multiple comparison tests} that aggregate the $p$-values $p_1,\ldots,p_k$ that come from the individual tests (in the case of treatment--control differences, typically individual $t$-tests). A standard aggregation procedure that is sometimes used to test the joint null hypothesis of no effect on a group of outcomes because of the simplicity of its implementation is the \textit{Bonferroni correction} \citep{bonferroni1936teoria,dunn1961multiple}: Only reject the null hypothesis of no effect at all if one of the $p$-values is smaller than $\alpha/k$, where $\alpha$ is the size of the test (typically $\alpha=5\%$) and $k$ is the number of outcomes.%
        \footnote{There is a related class of multiple testing approaches that controls the false discovery rate \citep{benjamini1995controlling}. Since our focus is on the group hypothesis of no overall effect, we do not further elaborate on this approach.}
        Improvements over these classical methods perform step-wise corrections that take into account the overall distribution of $p$-values \citep{holm1979simple} and/or the correlation between test statistics \citep{Romano:2005bw}.
        
        When would a procedure like the Bonferroni-corrected multiple-comparison test be appropriate?
        To understand how multiple-testing correction procedures perform in our application to the many-outcomes problem, it is important to understand that they are designed as an answer to a different question:
        when they reject the overall null hypothesis of no effect, they also provide rejections of individual hypotheses that are informative about which of the outcomes is affected by treatment.
        This additional information, however, comes at a cost; in particular with many outcomes, multiple-testing corrections tend to be conservative, and a direct test of the \textit{joint} hypothesis of no overall effect is a more efficient answer to the question we are asking in this paper.
        
        Beyond being limited to pre-specified means tests and paying a cost for providing individual rejections, Bonferroni-type corrections are inefficient in another way: Since they map individual (valid) $p$-values $p_1,\ldots,p_k$ from mean comparisons to an overall $p$-value without taking into account the correlations between test statistics, they are typically not both valid (that is, have size bounded by the nominal size) and optimal (that is, not be dominated by some other test across alternative hypotheses) without restrictions on the between-outcome covariance structure of the $\tau_j$ and $\epsilon_j$. In particular:
        \begin{enumerate}
            \item The Bonferroni correction (or its more powerful \cite{holm1979simple} stepwise version), which compares individual $p$-values against $\alpha/k$, is strictly conservative (has size smaller than nominal size) and inefficient under any correlation structure, and has maximal size only if $\tau_j$ and $\epsilon_j$ are independent between outcomes.
            \item The (more powerful) \cite{vsidak1967rectangular} correction, which compares individual $p$-values against $(1-\alpha)^{1/k}$, is valid, but strictly conservative unless $\tau_j$ and $\epsilon_j$ are independent between outcomes.%
            \footnote{The \v{S}id\'ak correction assumes non-negatively correlated $p$-values, which is given for two-sided $t$-tests under the null hypothesis.}
        \end{enumerate}
        Here, we assume that the equal-variance assumption from above holds, and that $p$-values are obtained from two-sided $t$-tests.
    
\begin{figure}
            \centering
            \begin{subfigure}{.5\textwidth}
              \centering
              \includegraphics[width=\textwidth]{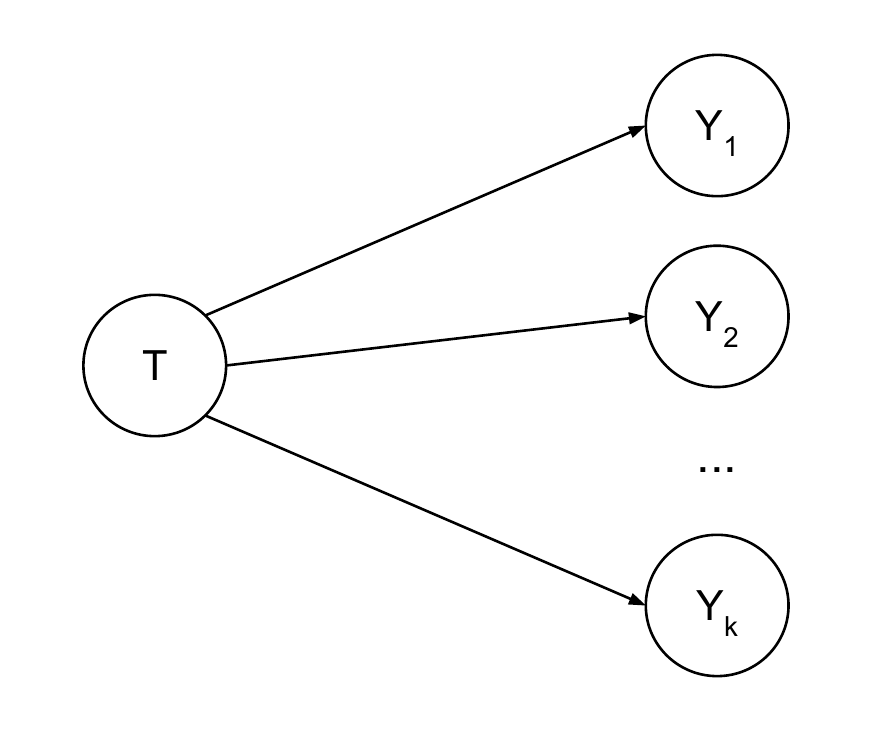}
            \end{subfigure}%
            \begin{subfigure}{.5\textwidth}
              \centering
                $Y_j=C_j+\tau_j T_j+\epsilon_j$
                
                $C_j$ constant
                
                $\tau_j$,$\tau_j$ independent of $\tau_{j'}$,$\tau_{j'}$ ($j' \neq j$)
            \end{subfigure}
            \caption{Conditional independence of outcomes}
            \label{fig:conditional_independence}
        \end{figure}

        Since such aggregation procedures do not use the information contained in the correlation of test statistics, they cannot generally be adequate, motivating variants that take the joint dependence structure of test statistics into account \citep{Romano:2005bw}.
        The two specific tests have largest size (although Bonferroni is still inefficient) for independent error terms and treatment effects between outcomes. This assumption means that outcomes are independent conditional on treatment assignment (only covary through the treatment), as expressed in the graphical model in Figure~\ref{fig:conditional_independence}.
    
        In applications in which all outcomes belong to a group, we typically assume them to be (conditionally) correlated, for example because they represent different aspects of one or more abstract (latent) outcomes that are affected by treatment.
        In this case,
        we may want to aggregate individual outcomes to one (or a few) aggregate outcomes on which we then perform means tests, rather than aggregating many tests with a multiple-testing correction.
        As an extreme case, assume we know how the outcomes are related, and that this relationship is through a single latent variable:

        \begin{rem}[A Case in which a Fixed Index is Appropriate]
            Assume that outcomes are correlated only through one common latent linear factor
            \begin{align*}
                L = K + \mu T + \delta
            \end{align*}
            the (Normal) distribution of which only differs between treatment and control by a mean shift, so in our Normal framework $\Var(\mu+\delta)=\Var(\delta)$ (that is, the data is as if the treatment effect $\mu$ on $L$ is constant). If we write $Y_j=C_j+\lambda_j L + \eta_j$ with fixed known treatment effect $\lambda_j$ and independent Normal mean-zero error terms with known variance $\nu^2_j$, then the linear index
            \begin{align*}
                I = \frac{\lambda_1}{\nu^2_1} Y_1 + \ldots + \frac{\lambda_k}{\nu^2_k} Y_k 
            \end{align*}
            is optimal in the sense that a Normal-theory means test on $I$ is an adequate test of the null hypothesis that treatment and control distributions are the same ($\mu=0$).
        \end{rem}
        
\begin{figure}
            \centering
            \begin{subfigure}{.5\textwidth}
              \centering
              \includegraphics[width=\textwidth]{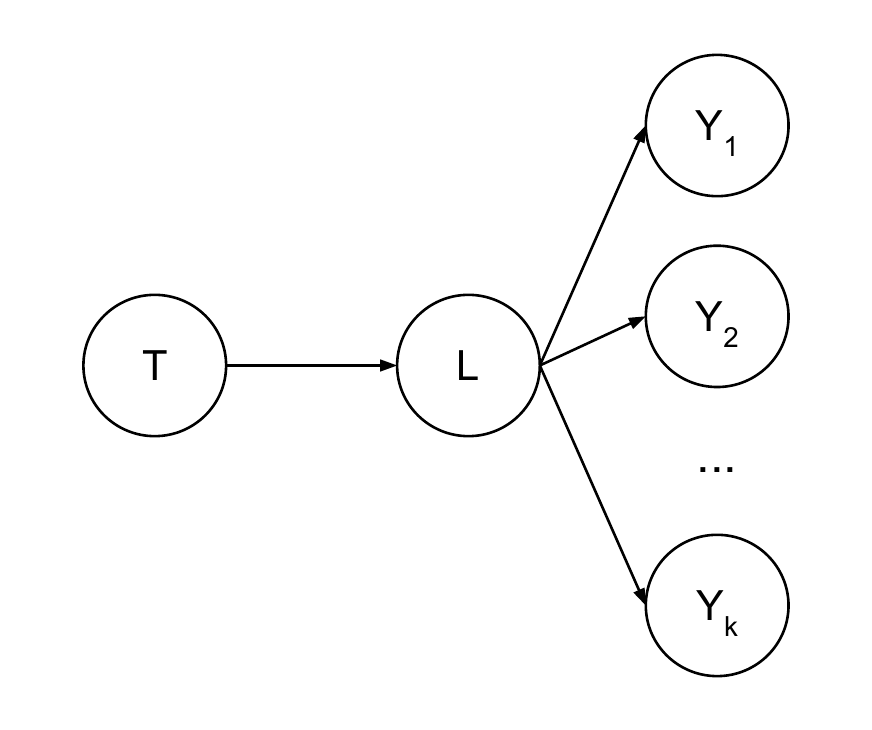}
            \end{subfigure}%
            \begin{subfigure}{.5\textwidth}
              \centering
                $L = K + \mu T + \delta$
                
                $Y_j = C_j + \lambda_j L + \eta_j$
                
                $K,C_j,\lambda_j$ constant
                
                $\eta_j$ independent of each other and of $\delta$
            \end{subfigure}
            \caption{One latent variable}
            \label{fig:one_latent_variable}
        \end{figure}

        If all correlation stems from one latent linear factor, and the treatment only operates through this factor, the outcomes are independent conditionally on this factor (Figure~\ref{fig:one_latent_variable}). In this case, an index (which estimates this latent factor) yields an adequate test by weighting each outcome by its signal–to–noise ratio.
        
        The above linear index aggregates outcomes efficiently into a single test statistic provided that:
        \begin{enumerate}
            \item The correlation structure is simple (in our example given by a single latent linear factor); and
            \item The relationships of outcomes to the latent factor are known ex-ante.
        \end{enumerate}
        So how can we aggregate across outcomes efficiently if we do not know the correlation structure? Joint tests like a Wald test in a SUR, which run all means tests simultaneously and produce a single $p$-value, can correct for unknown linear correlation in test statistics by estimating this correlation from the data; as long as the impact on the distribution is fully captured by the means, a linear SUR is thus adequate:

        \begin{rem}[Adequacy of Joint SUR Test]
            Assume that the data is generated from the Normal data-generating process introduced above. If the (multi-dimensional) distribution of treatment and control outcomes differs only by a mean shift, that is, if the variance–covariance matrix is the same,
            \begin{align*}
                \Var(\tau + \epsilon) = \Var(\epsilon)        
            \end{align*}
            then the Wald test in a linear SUR captures the difference between distributions. In other words, it is optimal (in an appropriate sense) if the distribution is observationally equivalent to one with constant (non-stochastic) treatment effects $\tau_j$.
        \end{rem}
    
        The intuition behind this result is straightforward: If the test comes from a regression that is correctly specified, it is adequate. If the two distributions, on the other hand, differ in more than their means, this specific regression is misspecified, and a test that also tests the effect on variances or correlations between outcomes may have higher power.

\begin{figure}
            \centering
            \begin{subfigure}{.5\textwidth}
              \centering
              \includegraphics[width=\textwidth]{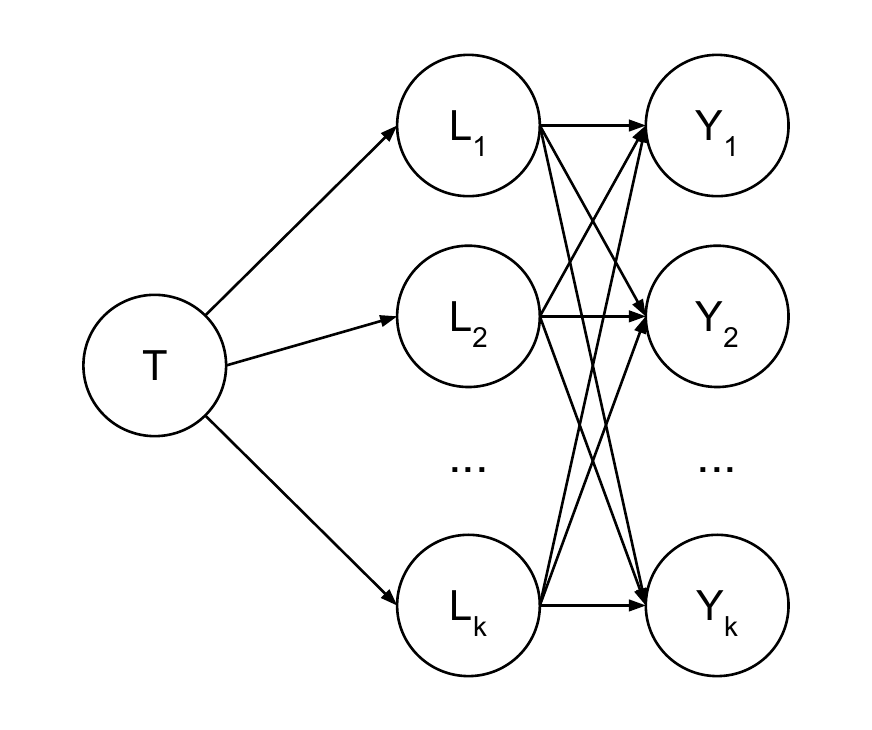}
            \end{subfigure}%
            \begin{subfigure}{.5\textwidth}
                \centering
                \begin{align*}
                    L_j &= K_j + \mu_j T + \delta_j \\
                    Y_j &= C_j + \sum_{j'=1}^k \lambda_{jj'} L_{j'} + \eta_j
                \end{align*}
                
                $K_j,C_j,\mu_j,\lambda_{jj'}$ constant
                
                $\delta_j,\eta_j$ all independent\footnotemark
            \end{subfigure}
            \caption{Conditional independence of latent variables}
            \label{fig:full_latent}
        \end{figure}

        Note that we do not require that individual outcomes are conditionally independent -- indeed, the SUR-based test can correct for correlation between the test statistics. However, the structural assumptions we put on the test to show efficiency imply that the outcomes are only affected by treatment through constant treatment effects on at most $k$ latent variables that are independent conditional on treatment (Figure~\ref{fig:full_latent}). We can thus think of the correction that SUR makes in constructing a test statistic as running the test not on the observed, correlated variables, but instead on the underlying (conditionally) uncorrelated latent factors $L_j$, each of which is affected by the treatment through a mean shift, and then aggregating these independent test statistics efficiently.
        
        \footnotetext{Leaving out/setting to zero all $K_j,\eta_j$ is without loss of generality.}

    \subsection{Economic Examples}

        So how do the structures assumed by standard approaches to testing effects on a group of outcomes map into economic applications of the sort encountered by experimental economists in actual data applications? We discuss several hypothetical models that are intended to be quite simple in their structure and yet sufficient to highlight the difficulty with which standard methods will be able to capture even this simple structure.

        \begin{ex}[Breaking a Causal Link]
            Assume that a health intervention tries to reduce the risk that high blood pressure leads to a heart attack- that is, it targets the link between hypertension and heart attack. Given data $(T_i,Y_{iA},Y_{iB})$ on treatment (assumed random), heart attacks, and blood pressure, how do we test whether the intervention has an effect?
            
            \begin{figure}[H]
            \centering
            \begin{subfigure}{.5\textwidth}
              \centering
              \includegraphics[width=.5\textwidth]{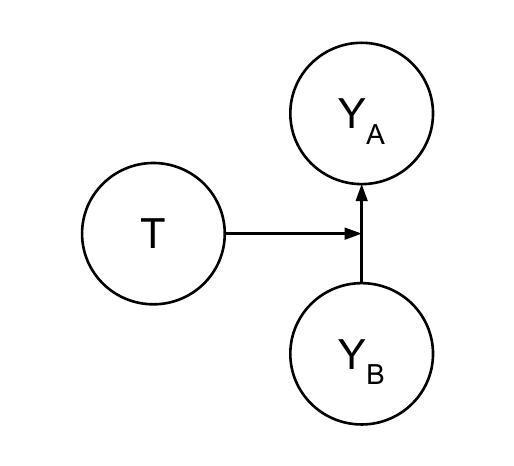}
            \end{subfigure}%
            \begin{subfigure}{.5\textwidth}
                \centering
                \begin{align*}
                    Y_A &= A_0 + \alpha Y_B + \beta T Y_B + \epsilon_A  \\
                \end{align*}
            \end{subfigure}
            \end{figure}

            In our model, treatment presumably affects heart attacks $Y_A$ positively, while not affecting blood pressure $Y_B$.
            
            If we test whether there is an overall effect of the intervention by running individual test on mean effects (multiple testing), and then aggregate the results using a Bonferroni correction to avoid false positives, including blood pressure data may reduce, rather than increase power, as it does not add additional independent information, but forces us to correct for the additional test. Even if there is some effect on blood pressure, a Bonferroni test would remain inefficient given the likely correlation between both test statistics.
            
            While a joint test like SUR would correct for these correlations, it remains limited on a more fundamental level because it only considers the impact on means. While not reducing power, $Y_B$ would at least not help SUR.
            
            But there \textit{is} meaningful information in blood pressure $Y_B$ that can help to increase the power of the test: Indeed, it is in the link between both variables where the effect of the intervention becomes most apparent – while in the control group those with higher blood pressure have more heart attacks, this link may be broken among the treated, as people with high blood pressure but without heart attack incidence represent the clearest evidence of an effect.
            
            % This is hardly the only type of data structure an experimental economist may encounter in the real world that would not fit very comfortably alongside the assumptions required by standard methods for testing effects on a group of outcomes.
        \end{ex}
        
        \begin{ex}[Creating a Correlation]
            Consider the effect of health insurance. Assume that there is some acute condition, say back pain, and some chronic condition, say some heart disease that leads to heart attack if not discovered and treated. If a person gets hospitalized because of back pain, we hypothesize that health insurance increases their propensity to then also get checked (and treated if necessary) for heart disease. Given data $(T_i,Y_{iA},Y_{iB})$ on treatment (assumed random), heart attacks, and back pain (which leads to hospitalizations $H_i$), how do we best test whether the intervention has an effect on health outcomes?

            \begin{figure}[H]
            \centering
            \begin{subfigure}{.5\textwidth}
              \centering
              \includegraphics[width=.5\textwidth]{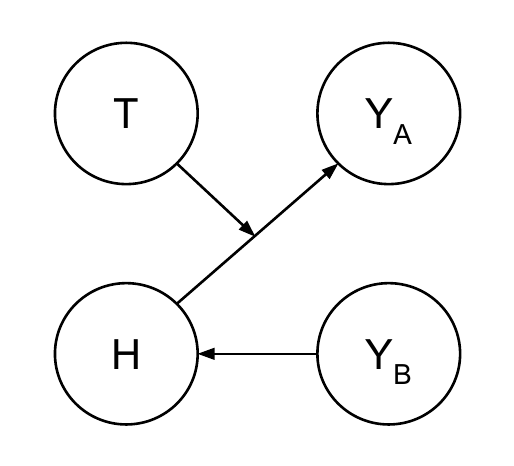}
            \end{subfigure}%
            \begin{subfigure}{.5\textwidth}
                \centering
                \begin{align*}
                    Y_A &= A_0 + \alpha T H + \epsilon_A \\
                    H &= H_0 + \beta Y_B + \epsilon_B
                \end{align*}
            \end{subfigure}
            \end{figure}
        \end{ex}

        \begin{ex}[Budget Constraint]
            Consider the effect of a cash grant, which we hypothesize increases the amount of money spent on education and health. However, every additional dollar spent on education or health can only be spent on one of these, moderated by the (unobserved) propensity of spending on one over the other.
            
            Given data $(T_i,Y_{iH},Y_{iE})$ on treatment (assumed random), health spending, and education spending (but not on the unobserved spending $M_i$ on health and education combined and individual propensity $P_i$ to spend on health versus education), how do we best test whether the intervention has an effect?

            \begin{figure}[H]
            \centering
            \begin{subfigure}{.5\textwidth}
              \centering
              \includegraphics[width=.5\textwidth]{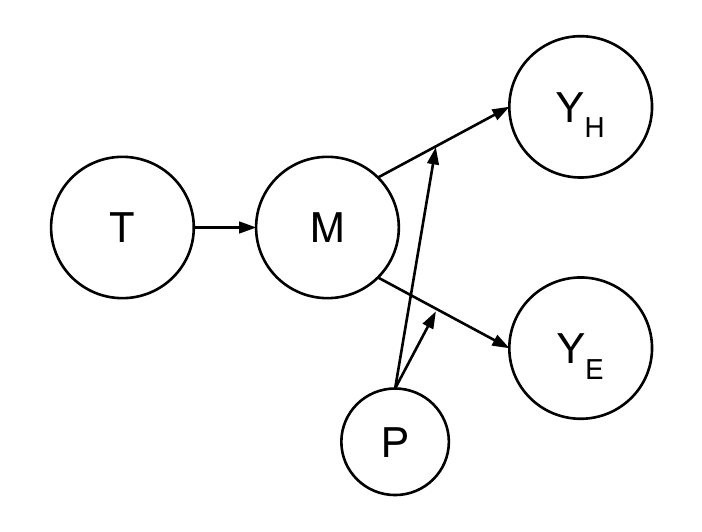}
            \end{subfigure}%
            \begin{subfigure}{.5\textwidth}
                \centering
                \begin{align*}
                    M   &= M_0 + \tau T + \epsilon_M \\
                    Y_H &= P M + \epsilon_H \\
                    Y_E &= (1-P) M + \epsilon_E
                \end{align*}
            \end{subfigure}
            \end{figure}
        \end{ex}
        
        \begin{ex}[On a Causal Chain]
            Consider the effect of an intervention that encourages preventive care on short- and long-run health outcomes. For example, statins (a preventive intervention) reduce cholesterol levels (a short-run outcome), which are associated with cardiovascular disease (a long-run outcome). Given data $(T_i,Y_{iC},Y_{iA})$ on treatment (assumed random), cholesterol levels (affected by statins take-up $S$), and heart attack, how do we best test whether neighborhoods have an effect?
        
            \begin{figure}[H]
            \centering
            \begin{subfigure}{.5\textwidth}
              \centering
              \includegraphics[width=.5\textwidth]{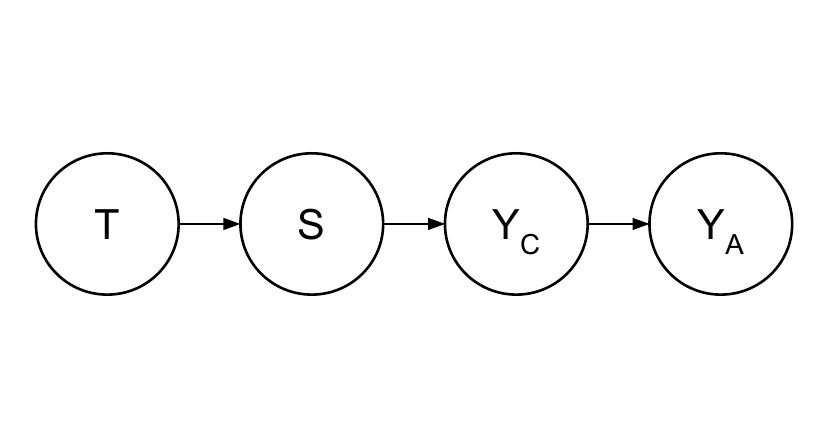}
            \end{subfigure}%
            \begin{subfigure}{.5\textwidth}
                \centering
                \begin{align*}
                    S &= S_0 + \tau T + \epsilon_S \\
                    Y_C &= C_0 + \gamma S + \epsilon_C \\
                    Y_A &= A_0 + \alpha Y_C + \epsilon_A
                \end{align*}
            \end{subfigure}
            \end{figure}
        \end{ex}

        \begin{ex}[Two Paths to Success]
            Consider the effect of moving to a new neighbourhood, which has potential effects both on parents’ earnings and school quality – directly through the availability of schools, and indirectly through the parents’ ability to afford a better school. Given data $(T_i,Y_{iE},Y_{iS})$ on treatment (assumed random), parents’ earnings, and education outcomes (as a proxy for unobserved school quality $Q_i$), how do we best test whether neighborhoods have an effect?

            \begin{figure}[H]
            \centering
            \begin{subfigure}{.5\textwidth}
              \centering
              \includegraphics[width=.5\textwidth]{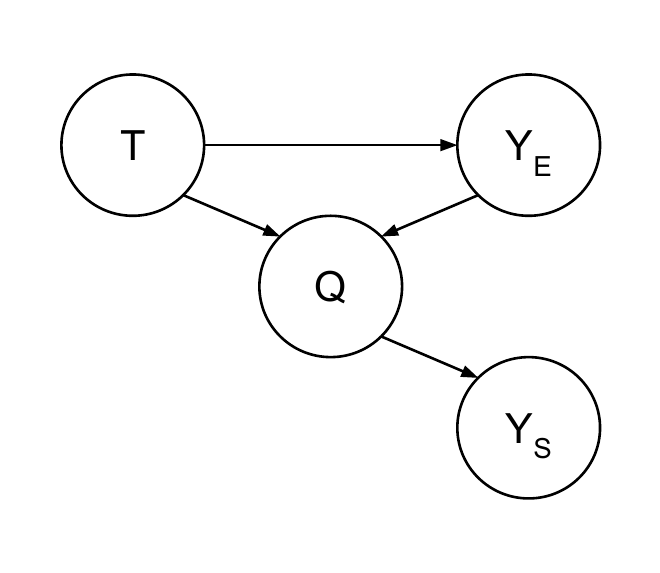}
            \end{subfigure}%
            \begin{subfigure}{.5\textwidth}
                \centering
                \begin{align*}
                    Y_E &= E_0 + \tau_E T + \epsilon_E \\
                    Q   &= Q_0 + \tau_Q T + \alpha Y_E + \epsilon_Q \\
                    Y_S &= S_0 + \beta Q + \epsilon_S
                \end{align*}
            \end{subfigure}
            \end{figure}
        \end{ex}
        
%        \q{ Alternatively/additionally: effect of earnings on outcomes conditional on quality? }
        
    \subsection{Take-Aways from Examples and Standard Approaches}

        In these examples, standard approaches leave information on the table, as they are misspecified for the economics of the situation. Each asks us to anticipate the effects we expect to see, but hardly any of the assumptions that justify the above tests maps well into the structure we have presented.
        
        There are two ways to go from here: First, we could try to build the right test for the right situation, using case-specific knowledge and theory to guide which structural assumptions we feel comfortable with each time and how we go about exploiting this structure by creating an appropriate set of individual hypotheses; or second, we could be agnostic about the specific structure and instead leverage tools –- such as those provided by machine learning –- that are able to adapt flexibly to any given dataset, and thereby explode the number of individual hypotheses they search over.
        We take this second path.

    \section{Prediction Test for an Effect on the Distribution}
    \label{sect:predicton}
        
        We turn the basic null hypothesis – ``the distribution of outcomes is the same in treatment as in control'' -- into a prediction statement: ``treatment status is not predictable from the outcomes''. We thus aim to predict $T$ from $Y$ using a flexible prediction function; if we find a function that predicts treatment status significantly better than chance, we have evidence of a difference between the two distributions.
        
        The workhorse of our testing procedure will thus be a prediction function: It takes as input a sample of treatment assignment and outcomes, and constructs from this sample a prediction function that maps values of the outcomes to the probability of being in the treatment group. The better this prediction is -- that is, the lower the discrepancy between the predicted probabilities and realized treatment assignment -- the better the evidence that treatment and control groups are different.
        
        One advantage of re-framing the group outcomes problem into one of prediction is that it enables us to take advantage of machine-learning approaches that are capable of considering very complicated functional forms. Moreover we can use the machinery of machine learning that guards against over-fitting in standard prediction applications to help find effects in a given experiment that reflect real underlying structure, rather than simply being artifacts of a particular sample.
        
        \subsection{Formal Framework}
        \label{sect:framework}
        
        To lay out our testing procedure,
        we make minimal assumptions on the data-generating process and the prediction function:
        \begin{itemize}
            \item \textbf{Random sample from RCT.}
                We have a sample
                \begin{align*}
                    S = \{ (T_1,Y_1),\ldots,(T_n,Y_n) \}
                \end{align*}
            of $n$ iid observations $(T_i,Y_i)$, with binary treatment $T_i$ assigned randomly (as in Section~\ref{sect:setup}).
            \item \textbf{Black-box prediction algorithm.}
                We have a prediction algorithm $A$ that takes as input a similar (but possibly smaller) sample $\tilde{S}$ of size $m \leq n$, and outputs a prediction function
                \begin{align*}
                    \fh = A(\tilde{S}): Y \mapsto \hat{T}=\fh(Y) \in \R
                \end{align*}
                While not formally required for our results until later, we assume that the function $\fh$
                is chosen to yield low out-of-sample loss
                \begin{align*}
                    L(f) = \E(\ell(f(Y),T))
                \end{align*}
                for some loss function $\ell: (\hat{T},T) \mapsto \ell(\hat{T},T) \in \R$; for most of this paper, we will assume that this is squared-error loss,
                \begin{align*}
                    \ell(\hat{T},T) = (\hat{T} - T)^2,
                \end{align*}
                but most results would go through for other well-behaved loss functions as well, such as the negative log-likelihood $\ell(\hat{T},T) = - (T \log(\hat{T}) + (1-T) \log(1 - \hat{T}))$.%
                \footnote{
                    Indeed, the negative log-likelihood offers an alternative justification for our approach, and shows its generality:
                    If instead we fitted a model $p_\theta(t,y) = p_\theta(t|y) p_\theta(y)$ of the distribution of $(T,Y)$ and assumed that
                    $p_\theta(t,y) = p_{\theta_1}(t|y) p_{\theta_2}(y)$ with $\Theta = \Theta_1 \times \Theta_2$,then the null hypothesis of equal conditional distributions, $p_\theta(y|t=1) = p_\theta(y|t=0)$ a.s., would translate into $p_{\theta_1}(t|y) = p_{\theta_1}(t)$ a.s.
                    Since the log-likelihood would factorize as
                    $\log p_\theta(t,y) = \log p_{\theta_1}(t|y) + \log p_{\theta_2}(y)$,
                    a conditional likelihood ratio test with $p_{\theta_1}(t|y)$ would be equivalent to a  likelihood ratio test with the full model.
                    This conditional likelihood ratio test is a prediction test with $\hat{f}(Y) = p_{\hat{\theta}_1}(1|Y)$ and the negative log-likelihood as loss.
                }
        \end{itemize}
        We will first introduce the testing procedure within this minimal framework,
        and then provide formal statements of its econometric properties.
        The validity of the test will not require any additional assumptions;
        for efficiency, we will assume more structure both about the data-generating process and the prediction technology.
        
        Note that we intentionally do not restrict the class of algorithms for now and do not even formally require that they \textit{minimize} out-of-sample loss $L(f)$ over some class of functions. This is to lay bare the relationship between prediction (of any quality) and the underlying hypothesis testing problem.  Practically, of course, the power of our testing procedure comes from the existence of machine-learning algorithms $A$ that produce low out-of-sample prediction loss $L(\fh)$. As a result, when we prove formal statements about the power of our procedure, we will make further assumptions.

        \subsection{Testing Procedure}
        
        The basis of our test is simple: compute out-of-sample loss $\hat{L}$ in predicting treatment $T$ from $Y$; compute the distribution of out-of-sample loss under the null of no treatment effect (where $T$ cannot be predicted from $Y$ at all); and form a test based on where $\hat{L}$ falls in this distribution. Since in-sample loss is a biased measure of out-of-sample performance (the function may look good in-sample through overfitting, even if there is no signal), and since we do not want to restrict prediction algorithms to those few for which we know how to estimate out-of-sample loss from in-sample loss, we rely on sample-splitting methods in evaluating loss: we never fit a function on the same data that we evaluate it on.
        
        The \textit{first variant} applies this idea in the most straightforward way:
        \begin{itemize}
            \item[\textbf{(H)}] \textbf{Hold-out test:} Split the sample randomly into a training sample $S_T$ of size $m$ and a hold-out sample $S_H$ of size $n-m$.
                Obtain a prediction function
                \begin{align*}
                    \fh_T = A(S_T)
                \end{align*}
                from the training sample.
                On the hold-out sample $S_H$, calculate predicted values $\hat{T}_i = \fh_T(Y_i)$ and
                test whether the loss
                \begin{align*}
                    \hat{L} = \frac{1}{n - m} \sum_{i \in S_H} \ell(\hat{T}_i,T_i) = \frac{1}{n - m} \sum_{i \in S_H} \ell(\fh_T(Y_i),T_i)
                \end{align*}
                is significantly better than if $T$ were unrelated to $Y$ (more on that below).
        \end{itemize}
        Note that this hold-out strategy ensures that all predictions are out-of-sample; however, it has an obvious inefficiency, as only a fraction of datapoints is used to fit the function -- and only a fraction is used to test its fit. This is corrected by our \textit{second variant}, which uses every datapoint once for evaluating loss, and multiple times for fitting:
        \begin{itemize}
            \item[\textbf{(CV)}] \textbf{Cross-validation test:} Split the sample randomly into $K$ equally-sized folds $S_1,\ldots,S_K$ (of size approximately $n/K$ each).
        For every $J \in \{1,\ldots,K\}$,
        obtain a prediction function
                \begin{align*}
                    \fh_J = A\left( \bigcup_{J' \neq J} S_{J'} \right)
                \end{align*}
                from all \textit{other} folds,
                and calculate predicted values $\hat{T}_i = \fh_J(Y_i)$ on fold $S_J$.
                Test whether the loss
                \begin{align*}
                    \hat{L} = \frac{1}{n} \sum_{i=1}^n \ell(\hat{T}_i,T_i) = \frac{1}{n} \sum_{J=1}^K \sum_{S_J} \ell(\fh_J(Y_i),T_i)
                \end{align*}
                is significantly better than if $T$ were unrelated to $Y$ on the full sample.%
                \footnote{Cross-validation is also used in tuning regularization parameters of many machine-learning algorithms. Here, we assume that any such tuning is happening ``inside'' the algorithm $A$, separately from our ``outside'' cross-validation loop. In the Appendix, we will discuss an implementation that combines the tuning of ensemble weights with the outer cross-validation to reduce computational cost.}
        \end{itemize}
        This procedure thus predicts $K$ times from $m = n - n/K$ datapoints;
        for $K=n$ (leave-one-out), the sample size of each training sample is maximal at $m=n-1$.
        
        \paragraph{Obtaining a $p$-value.}
        Given a loss estimate, how do we obtain a $p$-value for the null hypothesis that the distribution of outcomes is not affected by the treatment?
        
        We rely on sample-splitting methods and restrict ourselves to evaluating out-of-sample loss because we want to be agnostic about the algorithm used for prediction -- after all, we want to allow for complex machine-learning algorithms that may not fit our usual analytical estimation frameworks.
        For the same reason, we want to obtain a valid test under minimal assumptions.
        We therefore propose a permutation test:
        \begin{itemize}
            \item[\textbf{(H)}] For the hold-out design, permute treatment assignment randomly within the hold-out sample to obtain loss estimates
            \begin{align*}
                \hat{L}_\pi = \frac{1}{n - m} \sum_{i \in S_H} \ell(\hat{T}_i,T_{\pi(i)}).
            \end{align*}
            The $p$-value is the quantile of loss $\hat{L}$ within the distribution of $\hat{L}_\pi$.
            \item[\textbf{(CV)}] For the cross-validation design, permute treatment assignment randomly across the full sample, re-estimate all prediction functions, and obtain loss estimates $\hat{L}_\pi$.
            The $p$-value is the quantile of loss $\hat{L}$ within the distribution of $\hat{L}_\pi$.
        \end{itemize}
        The logic behind the permutation test is straightforward:
        If the null hypothesis is true, the data with permuted treatment assignment looks just like the original sample;
        if, to the contrary, we can predict better from the actual sample, this is evidence for a treatment effect.
        Crucially, this logic does not depend on any specific feature of the data-generating process or the algorithm, but it reflects the strength of the null hypothesis.
        
        Note that the hold-out test does not require the re-estimation of prediction functions in every run; the cross-validation scheme, on the other hand, refits $K$ prediction function for every permutation draw.\footnote{In the Appendix, we will discuss a hybrid approach that permutes both training and hold-out set in the hold-out design as a compromise between computational cost and statistical power.}
        We see the main disadvantage of the cross-validation procedure over the holdout procedure as coming from the fact that it is much more computationally costly -- both because of repeated predictions for one sample and re-estimation for every permutation -- although this cost can be mitigated to some degree by the fact that all predictions can be computed in parallel.

        \subsection{Econometric Analysis}

        We start with an analysis of our testing procedure under the minimal assumptions from Section~\ref{sect:framework}.
        By construction, the hold-out and $K$-fold permutation test have exact size (provided appropriate interpolation to form the $p$-value):
        
        \begin{clm}[Exact Size of H and CV tests]
        \label{prop:exactsize}
            The hold-out test and the $K$-fold permutation test that reject the null hypothesis (that treatment assignment is independent of the outcomes) whenever $p<\alpha$ has (exactly) size $\alpha$, for any nominal level $\alpha$ and sample size $n$.
        \end{clm}
        
        Note that we did not make any assumptions about the prediction algorithm; in particular, the tests have exact size even for complex machine-learning prediction algorithms.

        When deciding whether to employ the above testing procedure, we care not just about size, but also power. In particular, we may be worried that we lose efficiency relative to a standard joint test based on mean differences in cases in which the effect is exclusively on the means of the outcomes, or relatively to a prediction-based test that uses the full sample and corrects for overfitting analytically. Towards efficiency, we first show asymptotic equivalence if we restrict ourselves to linear predictors,
        where the data comes from a linear model with Normal, homoscedastic errors:
        
        \begin{asm}[All Effects are Mean Shifts]
        \label{asm:meanshifts}
            Assume that the data is generated by
            \begin{align*}
                Y=C+\tau_n T+\epsilon
            \end{align*}
            where
            \begin{enumerate}[label=(\ref{asm:meanshifts}.\arabic*)]
                \item $T$ independent of $\epsilon$,
                \item $T$ has fixed distribution (across $n$) with $p = \E T \in (0,1)$,
                \item fixed dimension $k$ of $Y$, $C$, $\tau_n$, $\epsilon$,
                \item $\epsilon \sim \N(\0,\Sigma)$  with $\Sigma$ symmetric and positive-definite,
                \item $C$ is a constant vector,
                \item treatment effects are local to zero with $\tau_n=\tau_1/ \sqrt{n}$,
                where $\tau_1$ is a constant vector.
                \label{asm:meanshifts-local}
            \end{enumerate}
        \end{asm}

        We formulate these specific conditions because they represent a world in which a linear SUR is correctly specified and the corresponding Wald test is a natural test for the joint hypothesis of no effect.
        Indeed, these assumptions ensure that the full treatment effect is captured by mean shifts in the individual outcome dimensions.
        The normalization \ref{asm:meanshifts-local} provides that asymptotic power does not take off to one asymptotically.
        We claim that the following equivalence results holds under these conditions:
                
        \begin{clm}[Asymptotic Equivalence of Linear Tests]
        \label{prop:equivalence}

            Under \autoref{asm:meanshifts}, the following tests are asymptotically equivalent in the sense that their power converges to the same limit:
            \begin{enumerate}
                \item A joint $\chi^2$-test (Wald test) in the linear regression (SUR) of $Y$ on $T$;
                \item An $F$- (goodness-of-fit) or $\chi^2$-test (Wald test) in an in-sample linear regression of $T$ on $Y$;
                \item The $K$-fold permutation test with linear least-squares prediction function and squared-error loss for $K=n$ (leave-one-out).
            \end{enumerate}
        \end{clm}
        
        A precise statement of this and all following results will be found in the Appendix.
        
        Hence, as long as we use the linear-least squares estimator as a predictor, we do not asymptotically lose any power from putting the outcomes on the right-hand side of the regression, or from running a cross-validation permutation test.
        In particular, there is no (asymptotic) loss from using the prediction test with a linear predictor relative to a standard test based on pairwise comparisons.

While these results are reassuring, we motivated our prediction procedure through more flexible functional forms.
So what happens if OLS is \textit{not} correctly specified, and a more flexible algorithm will find additional structure?

        \begin{clm}[Consistency and Asymptotic Dominance of Better Predictor]
        \label{prop:dominance}
            If the algorithm $A$ recovers more information in the limit than a trivial predictor in the sense that
            \begin{align*}
                \limsup_{n \rightarrow \infty} \E_n(L(\fh)) < \inf_{t \in \R} L(t),
            \end{align*}
            then, if the prediction algorithm is well-behaved in the sense that
            \begin{align*}
                \E_n \Var_n(\fh(Y)|Y) \rightarrow 0,
            \end{align*}
            and under appropriate regularity conditions on the sequence of data-generating processes,
            the $K$-fold and hold-out tests -- using the algorithm $A$ directly or inside an ensemble -- have power approaching one (i.e. is consistent). 
            In particular, unless the Wald test in linear regression also has power converging to one (or OLS as a predictor recovers more predictive information in the limit than a trivial predictor),
            the test based on the more predictive algorithm has power asymptotically strictly larger than the linear Wald test (or a prediction test with linear regression).
        \end{clm}
        
        This result is merely a formalization of the intuition we started our inquiry with:
        If our predictor picks up structure that is not linear, and that corresponds to a definite loss improvement in the large-sample limit, then it will ultimately reject the null hypothesis of no treatment effect,
        while a linear-based test may achieve strictly smaller power.

    \section{Interpreting the Prediction Output}
    \label{sect:analyze}
    
    The above test gives us a $p$-value for the null hypothesis that there is no difference in outcome distribution between treatment and control groups, based on a prediction task. However, we may also be interested in what the effect is on -- that is, where in the outcome space the prediction algorithm found signal about treatment. This is relevant for using experiments to test economic theories, which often generate testable implications about which outcomes should and should not be affected in an experiment, and for carrying out benefit--cost analyses of specific interventions. However, with machine-learning techniques this is more complicated than just testing for an overall effect on a group of outcomes -- that is, testing whether treatment assignment is predictable. The reason is that machine-learning tools are designed to generate good out-of-sample predictions by extracting as much signal as possible from the explanatory variables, rather than to isolate the  individual relationships of each predictor with the left-hand side variable. In this section we discuss how we navigate this constraint.
    
    \subsection{Index Interpretation}
    
    A prediction function of treatment assignment from the outcomes is itself a function $\fh(Y)$ of the outcomes $Y$, and thus an index. Indeed, we can interpret the prediction exercise with respect to squared-error loss as the estimation of an optimal index:
    
    \begin{rem}[Optimal Index Interpretation]
    \label{rem:optindex}
        The prediction function that minimizes squared-error loss within a given class of functions is also the function within that class with the maximal mean treatment–control difference (for a given mean and variance).
    \end{rem}
    
    When we choose a prediction function, we thus also estimate an optimal index -- where ``optimal'' refers to it being maximally different between treatment and control. This mean difference also offers a quantification of the difference between the distributions.

    \subsection{Analyzing the Difference Between the Two Distributions}
    
    If we are not just interested in whether there is an effect at all (expressed by said $p$-value), but also where it is, we can use the predicted treatment assignment values $\hat{T}_i$ obtained within the hold-out or $K$-fold testing procedure as a guide to where the two distributions differ;
    in particular, for standard loss functions, $\hat{T}$ can be seen as estimating $\P(T=1|Y)$, as $\P(T=1|Y)$ is the prediction minimizing squared-error loss and maximizing the likelihood.

    To obtain summary statistics of the difference in distributions expressed by these predicted values, we can calculate the implied treatment effect on any function of the outcome vector (including a given index), such as the outcomes vector itself:
    \begin{align*}
        \hat{\tau}_{\hat{T}}
        = \frac{1}{n}
        \sum_{i=1}^n \frac{\hat{T}_i - \frac{1}{n} \sum_{\ell=1}^n \hat{T}_\ell}{\left(\frac{1}{n} \sum_{\ell=1}^n \hat{T}_\ell \right) \left(1 - \frac{1}{n} \sum_{\ell=1}^n \hat{T}_\ell \right)} Y_i
    \end{align*}
    While primarily an expression of the treatment effect expressed by the prediction function, it becomes an estimate of the average treatment effect vector $\tau =\E(Y|T=1) - \E(Y|T=0)$ provided that the prediction function is loss-consistent (where loss is assumed to be out-of-sample squared-error loss of the algorithm throughout):

    \begin{clm}[Loss Consistency Implies Estimation Consistency]
    \label{prop:hatconsistency}
        If the predictor is loss-consistent,
        \begin{align*}
            L(\fh) \stackrel{\P_n}{\longrightarrow} L^*,
        \end{align*}
        where $L^*=L(f^*)$ is the loss from the optimal predictor $f^*(Y)=E(T|Y)$, then, under regularity conditions,
        \begin{align*}
            \hat{\tau}_{\hat{T}}  \stackrel{\P_n}{\longrightarrow} \tau.
        \end{align*}
    \end{clm}

    Note that this result can be extended to the effect on any fixed function $g(Y)$ of the outcomes.
    Similar results can be  achieved for expectations of functions $h(Y,\P(T=1|Y))$ that are estimated by the average of $h(Y,\hat{T})$.

    \section{Simple Representations of the Difference Between Distributions}
    \label{sect:difference}
    
    In this section, we propose one approach that directly optimizes for a simple representation of the causal effect on the outcome distributions.
    Specifically, we consider a discretization of treatment and control distributions stemming from a partition of the outcomes space.
    In maximizing the expressiveness of this simple representation, we link its construction back to the reverse regression problem our tests are based on.
    
    \subsection{Setup and Goal}
    
    We assume for simplicity that the outcome vector $Y$ is continuously distributed (with overall density $f(y)$) both in treatment (density $f_1(y)$) and control (density $f_0(y)$), and that $Y$ takes values in $\mathcal{Y}$.
    Our goal is to find a partition
    \begin{align*}
        \mathcal{Y} = \bigcup_{\ell=1}^L \mathcal{P}_\ell
    \end{align*}
    such that the discretized distributions
    \begin{align*}
        \bar{f}_1(\ell) &= \P(Y \in \mathcal{P}_\ell|T=1)
        &
        \bar{f}_0(\ell) &= \P(Y \in \mathcal{P}_\ell|T=0)
        &
        \bar{f}(\ell) &= \P(Y \in \mathcal{P}_\ell)
    \end{align*}
    preserve as much information about the difference between treatment and control as possible.
    For example, we consider as a criterion for the normalized differences
    $\Delta(\ell) = \frac{\bar{f}_1(\ell) - \bar{f}_0(\ell)}{\bar{f}(\ell)}$
    the variance
    \begin{align}
    \label{eqn:varmax}
        \Var_{\bar{f}}(\Delta) = \sum_{\ell=1}^L \bar{f}(\ell) \Delta^2(\ell),
    \end{align}
    which we want to maximize to obtain a discretization that is as expressive as possible about the causal effect on the distribution of outcomes.%
    \footnote{We present one approach, based on variance and leading to regression. Alternatively, expressing informatin about the distribution in terms of entropy/divergence could yield maximum-likelihood classification.}

    \subsection{Prediction Level Sets}
    
    The difference measure $\Delta$ connects the exercise of providing a representation of the difference between two distributions
    to the reverse regression of predicting $T$ from $Y$ since
    \begin{align*}
        \Delta(\ell) = \frac{\bar{f}_1(\ell) - \bar{f}_0(\ell)}{\bar{f}(\ell)}
        = \frac{\P(T=1|Y \in \mathcal{P}_\ell)}{\P(T=1)} - \frac{\P(T=0|Y \in \mathcal{P}_\ell)}{\P(T=0)}
        = \frac{\E(p(Y)|Y \in \mathcal{P}_\ell) - p}{p (1-p)}
    \end{align*}
    for $p = \P(T=1), p(y) = \P(T=1|Y=y) = \E(T|Y=y)$.
    In particular, if we do not further restrict the partitions (other than in terms of number of parts ans possibly their size), optimal sets are obtained as levels sets of $p(y)$:
    
    \begin{clm}[Unconstrained Optimal Partition]
        \label{prop:partitionsgeneral}
        If partitions of size $L$ are unconstrained (except possibly for the sizes $\bar{f}(\ell)$), then (for $p(Y)$ continuously distributed on $[0,1]$) an optimal partition is obtained from cutoffs 
        \begin{align*}
            0 = c_0 < c_1 < \ldots < c_L = 1+\underbrace{\varepsilon}_{> 0}
        \end{align*}
        (that may depend on the distribution) and
        \begin{align*}
            \mathcal{P}_\ell = \{ y \in \mathcal{Y}; c_{\ell - 1} \leq p(y) < c_\ell \}.
        \end{align*}
    \end{clm}
    
    This claim characterizes an optimal partition in terms of level sets of the oracle predictor $p(y) = \E(T|Y=y)$.
    A natural empirical implementation leverages predictions $\hat{T} = \hat{f}(Y)$, where $\hat{f}$ is formed on a training dataset $S_T$ by predicting treatment $T$ from outcomes $Y$, yielding a partition
    \begin{align*}
        \hat{\mathcal{P}} = \{y \in \mathcal{Y}; \hat{c}_{\ell-1} \leq \hat{f}(y) < \hat{c}_{\ell} \}.
    \end{align*}
    Here, the $\hat{c}_{\ell}$ could be appropriate quantiles of $\hat{T}$ obtained from the training dataset (e.g. by cross-validation).
    Specifically, we obtain a partition
    \begin{align*}
        \hat{\mathcal{P}}_H = \{i \in S_H;  \hat{c}_{\ell-1} \leq \hat{f}(Y_i) < \hat{c}_{\ell} \}
    \end{align*}
    of the units in the holdout, which allows for honest estimation \citep{Athey2016-zi} of the difference between these groups. Specifically, we can test whether treatment and control counts across subsets provide significant evidence of a difference in distributions as represented by the partition, and obtain unbiased estimates of the mean differences of individual outcome variables between groups.

    \subsection{Recursive Partitioning}
    
    While level sets based on predicted treatment provide a solution to the problem of partitioning the outcome space to represent the causal effect on the distribution,
    these level sets may have complex shapes that may at best be represented graphically for relatively low-dimensional outcome vectors, but are in general hard to describe.
    Following the work of \cite{Athey2016-zi} on heterogeneous treatment effects and \cite{Gagnon-Bartsch2019-ay} on analyzing baseline imbalance, we also consider further restricting the partitions to recursively defined axis-aligned rectangles (i.e., decision trees), assuming that $\mathcal{Y} \subseteq \R^k$.
    Restricting to partitions that can be represented as the leaves of a decision tree that splits on outcome variables, the problem of finding an optimal partition is infeasible even for the in-sample analog. However, maximizing the variance in $\Delta$ is equivalent to minimizing the prediction error of predicting $T$ from its average within subsets:
    
    \begin{clm}[Equivalence to Regression]
        \label{prop:partitionregressionequivalence}
        Finding a partition (among a set of candidate partitions) maximizing (\ref{eqn:varmax}) is equivalent to finding a partition minimizing regression loss
        \begin{align*}
            \E((T - \bar{p}(\ell(Y)))^2),
        \end{align*}
        where $\ell(Y)$ is the set $\ell$ of the partition with $Y \in \mathcal{P}_\ell$, and $\bar{p}(\ell) = \E(T|Y \in \mathcal{P}_\ell)$.
    \end{clm}
    
    As a consequence, we can apply standard regression trees that minimize mean-squared error in the prediction of $T$ from $Y$ to obtain a partition of $\mathcal{Y}$ into axis-aligned rectangles defined by the resulting decision tree.
    Running this prediction exercise in the training sample, we again obtain a partition of units in the hold-out (given by the leaves of the tree) that allows for honest estimation of conditional treatment probabilities as well as outcome mean differences between groups.

    \section{Simulation Example}
    \label{sect:simulation}
    
    As an illustration for the basic prediction test and to calibrate power, we provide a simulation example with $k=2$ outcomes based on the data-generating process
    \begin{align*}
        Y = \begin{pmatrix} 1 \\ 1 \end{pmatrix} T m + \epsilon (1 + T s) \in \R^2
    \end{align*}
    with $\epsilon$ two-dimensional standard Normal,
    where $m$ (``move'') and $s$ (``stretch'') are both scalar parameters: $m$ moves the treated distribution towards the north-east (relative to the control distribution, which is centered at the origin) and $s$ stretches the variance of the treated distribution equally in all directions (relative to the control distribution, which has rotation-invariant unit variance).
    We assume that $\P(T=1) = .5$ throughout.
    In this setting, the null hypothesis of no effect (same distributions) corresponds to $m= 0 = s$.
    For different values of the parameters, we study the power of a test of this null hypothesis from samples of size $n=100$.
    
    We compare two test for no overall effect of treatment:
    \begin{itemize}
        \item A Wald test based on a linear regression (SUR) of $(Y_1,Y_2)$ on $T$; this test is based on a correctly specified regression for no stretch ($s=0$), but cannot detect changes in the variance between treatment and control ($s \neq 0$).
        \item A 5-fold prediction permutation test as described above, with an ensemble of linear regression ($T$ on $Y_1,Y_2$ and constant) and a random forest of 100 trees, tuned using 5-fold cross-validation along the number of variables considered in tree splits and the minimal number of units in a node; the weights for the ensemble are chosen in-sample to minimize the loss of a convex combination of out-of-sample predictions from both predictors.
    \end{itemize}
    
    \begin{figure}
        \centering
        \begin{subfigure}{.45 \textwidth}
            \centering
            \includegraphics[width=\textwidth]{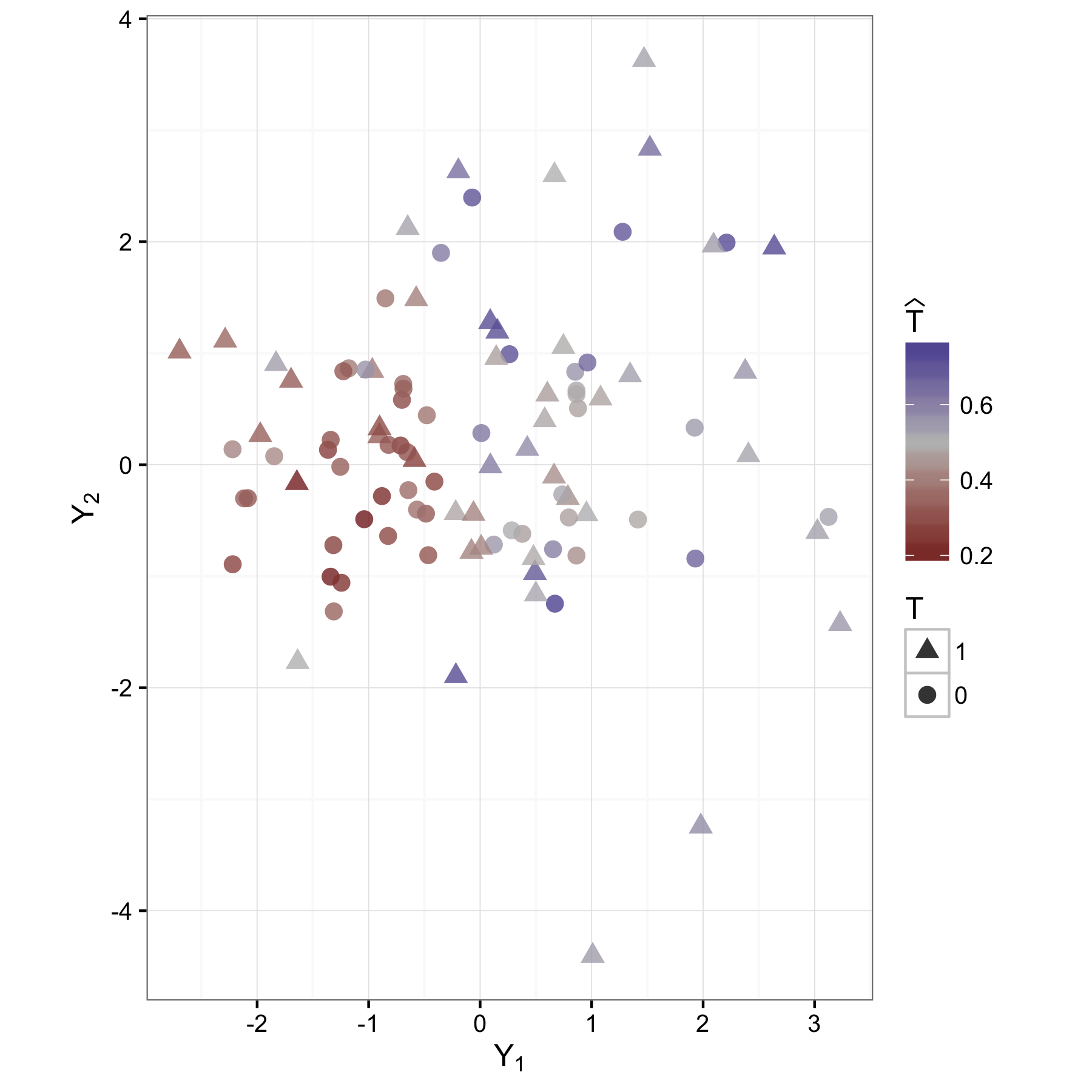}
            \caption{Visualization of data and predicted treatment assignment from a 5-fold cross-validated ensemble of linear regression and tuned random forest}
            \label{fig:simexamplefitted}
        \end{subfigure}
        \quad
        \begin{subfigure}{.45 \textwidth}
            \centering
            \includegraphics[width=\textwidth]{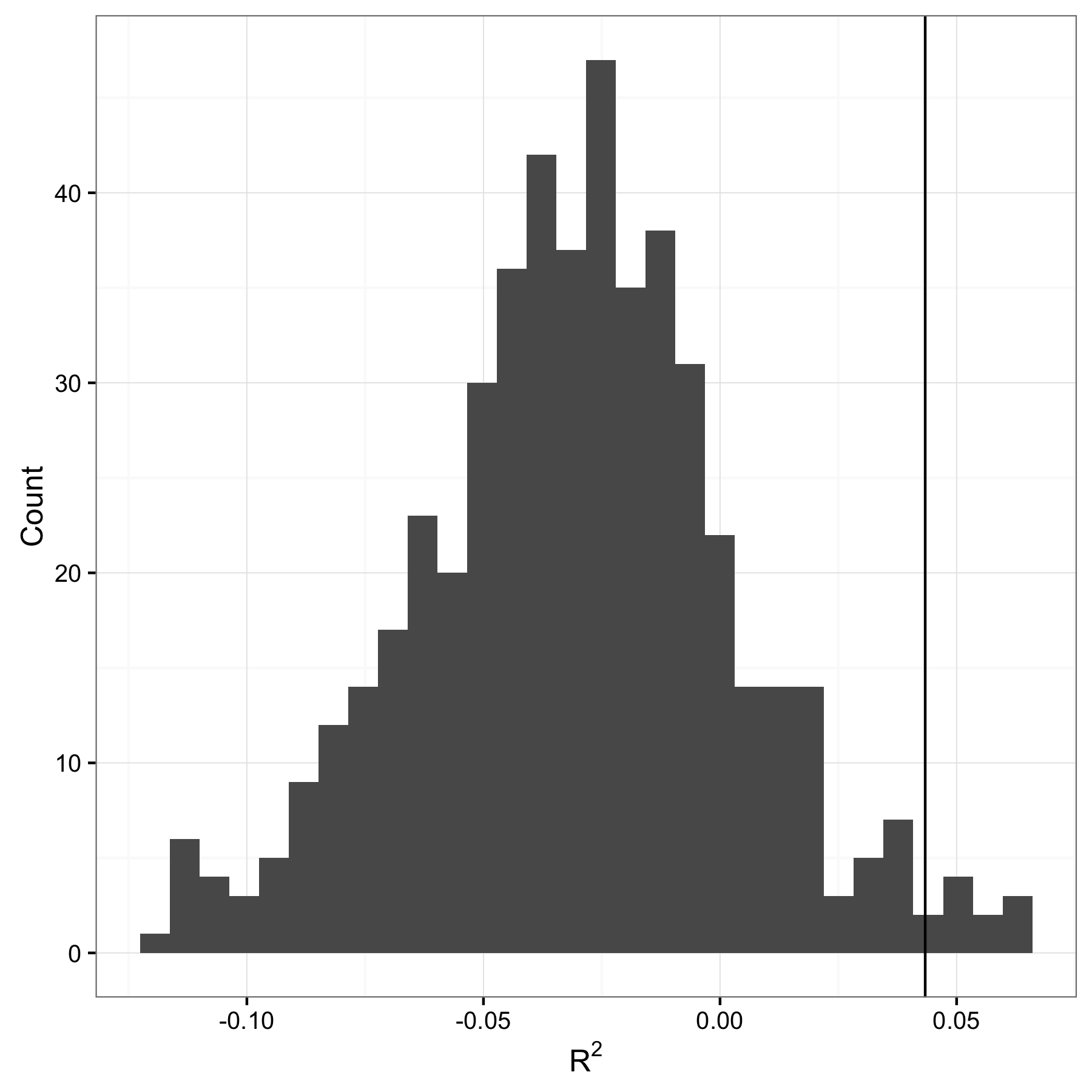}
            \caption{Histogram of fit for permutation draws (cross-validated $R^2$), with prediction test loss as horizontal line; the $p$-value is $2.0\%$, while a Wald test in a SUR yields a $p$-value of $10.0\%$}
            \label{fig:simexamplepermutation}
        \end{subfigure}
        \caption{Simulation example with $m = 0.2$, $s = 0.5$}
        \label{fig:simexample}
    \end{figure}

    As an illustration, \autoref{fig:simexample} presents the data and fitted ensemble predictions for one draw from this data-generating process with moderate move ($m=0.2$) and considerable stretch ($s=0.5$).
    For this draw, the predicted values $\hat{T}_i$ correctly locate the lower density of treatment in the lower-left quadrant and the higher density in the periphery (\autoref{fig:simexamplefitted}); the permutation test produces a $p$-value of $2\%$ (\autoref{fig:simexamplepermutation}) compared to a $p$-value of $10\%$ for the Wald test in a SUR on the same data.
    
    \begin{table}
    
        \begin{subtable}{\textwidth}
        \centering
        \begin{tabular}{lr rrrrrr}
        \toprule
        $s$ (stretch) & & 0.0 & 0.1 & 0.2 & 0.3 & 0.4 & 0.5 \\
        \midrule
        $m$ (move) & 0.0 &
         4.3\% & 6.2\% & 3.9\% & 5.9\% & 4.3\% & 5.5\%\\
         & 0.1 &
         9.3\% & 10.3\% & 6.8\% & 5.7\% & 9.8\% & 6.6\%\\
         & 0.2 &
        25.1\% & 19.6\% & 20.0\% & 17.1\% & 20.3\% & 15.0\%\\
         & 0.3 &
        45.6\% & 41.5\% & 40.8\% & 35.3\% & 32.6\% & 29.4\%\\
         & 0.4 &
        69.2\% & 66.5\% & 62.9\% & 59.7\% & 55.4\% & 52.8\%\\
         & 0.5 &
        90.0\% & 85.4\% & 80.6\% & 77.7\% & 72.0\% & 68.1\%\\
        \bottomrule
        \end{tabular}
        
        \caption{Wald test in a linear SUR of $Y_1,Y_2$ on $T$}
        \end{subtable}
        
        \medskip
        
        \begin{subtable}{\textwidth} 
        \centering
        \begin{tabular}{lr rrrrrr}
        \toprule
        $s$ (stretch) & & 0.0 & 0.1 & 0.2 & 0.3 & 0.4 & 0.5 \\
        \midrule
        $m$ (move)  & 0.0 &
        3.6\% & 5.0\% & 7.5\% & 17.5\% & 28.0\% & 42.4\%\\
        & 0.1 &
        8.9\% & 9.1\% & 10.9\% & 18.7\% & 30.8\% & 43.1\%\\
         & 0.2 &
        20.0\% & 16.2\% & 19.8\% & 28.9\% & 37.4\% & 49.4\%\\
         & 0.3 &
        34.6\% & 33.0\% & 34.6\% & 39.6\% & 49.2\% & 57.6\%\\
         & 0.4 &
        60.8\% & 55.4\% & 56.7\% & 60.6\% & 63.8\% & 68.6\%\\
         & 0.5 &
        79.3\% & 75.4\% & 75.4\% & 73.1\% & 78.6\% & 81.8\%\\
        \bottomrule
        \end{tabular}
        
        \caption{Prediction test based on ensemble of linear regression and random forest}
        \end{subtable}
                
        \caption{Rejection frequencies for varying values of move and stretch parameters; preliminary results based on 439 Monte Carlo draws}
    
        \label{tbl:simpower}

    \end{table}
    
    \autoref{tbl:simpower} reports the rejection frequency for the Wald and prediction test at a nominal level $\alpha = 5\%$ based on 439 Monte Carlo draws.
    Under the null hypothesis ($m = 0 = s$), the rejection frequencies (size) are at or below the nominal level for both tests.
    As move increases, the power of the Wald test and the prediction test increase, with the Wald test attaining higher power if there is no stretch.
    The picture is quite different for varying values of the stretch parameter:
    In the framework of linear regression of $Y_1,Y_2$ on $T$, higher stretch only adds variance and thus decreases power; for the prediction test, on the other hand, differential variance between control and treatment provides evidence of an effect and thus increases the rejection rate. 
    
    \begin{figure}
        \centering
        
        \includegraphics[width=\textwidth]{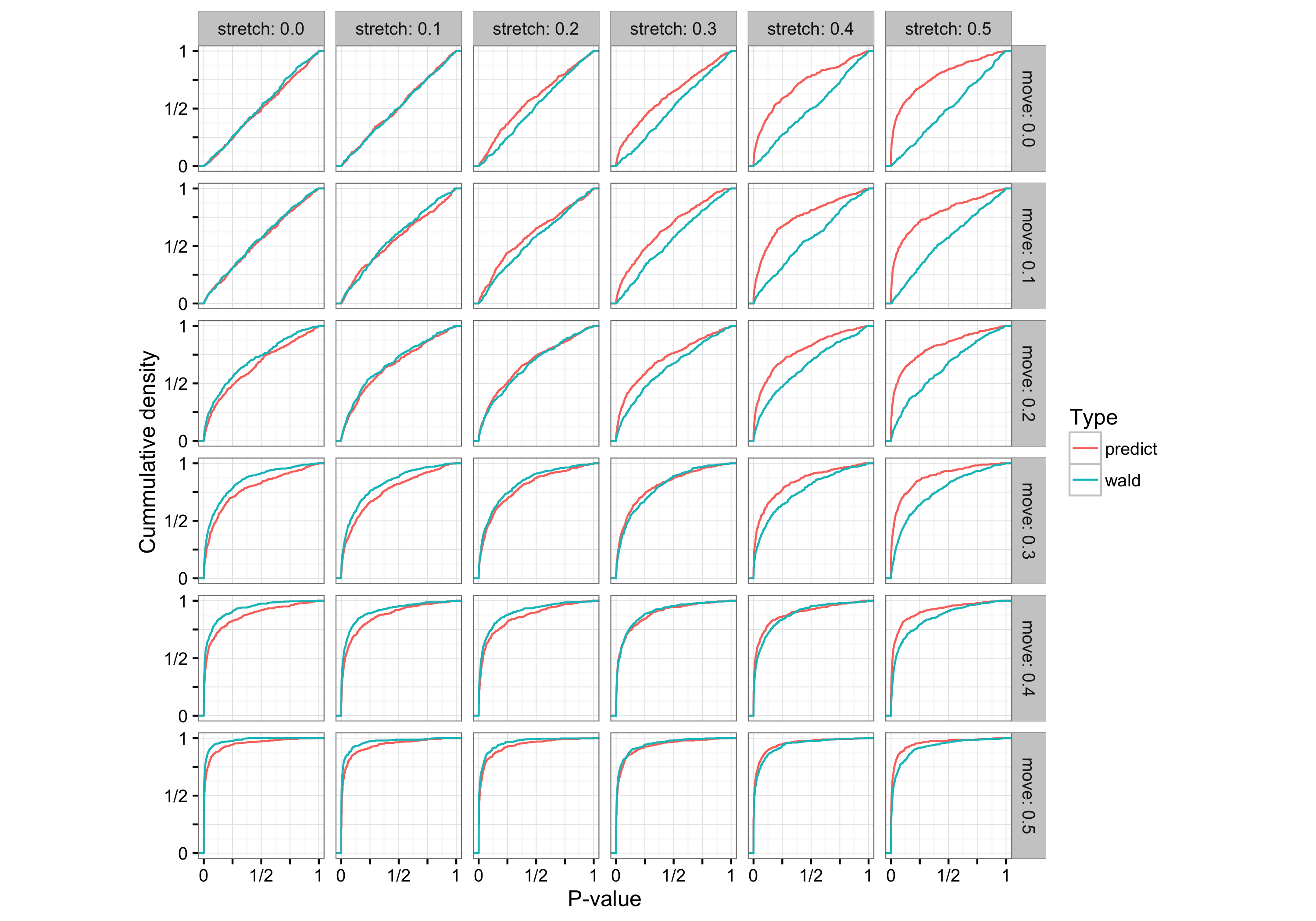}
    
        \caption{Empirical cumulative distribution functions of $p$-values for Wald and prediction test, for varying values of move and stretch parameters; preliminary results based on 439 Monte Carlo draws}
        \label{fig:simecdf}
    \end{figure}
    
    The results in \autoref{tbl:simpower} are restricted to a nominal level of $\alpha = 5\%$;
    \autoref{fig:simecdf} extends the analysis to the full empirical distribution of $p$-values (where the results in \autoref{tbl:simpower} can be obtained by evaluating each at $5\%$).
    Under the null hypothesis (zero stretch, zero move), the $p$-values are approximately uniform within the unit interval, representing performance close to nominal.
    The effect of move and stretch are as below, with move improving power of both tests, while stretch induces an increasing gap between the power of prediction and linear Wald tests represented by the wedge between the empirical cumulative distribution curves.   
    
    \section{Extensions}
    \label{sect:conclusion}
    
    We have presented a specific methodology that allows for leveraging powerful machine-learning predictors to test for the effect of a randomized intervention on a group of outcome variables.
    We note that our methodology can be adopted for outcome tests that include control variables (by including them in the prediction exercise), cluster randomization (by splitting and permuting data by clusters), stratified/conditional (within-site) randomization (by splitting and permuting within strata/sites), and missing data (e.g. by testing whether the outcomes predict treatment assignment better than missingness information alone).

    \bibliography{Bibliography}

\end{document}